\documentclass[letterpaper]{article} 
\usepackage[preprint]{aaai2027}    

\usepackage{amsmath}
\usepackage[hyphens]{url}            
\usepackage{graphicx}                
\usepackage{natbib}                  
\usepackage{caption}                 
\usepackage{booktabs}
\usepackage{subcaption}

\usepackage{hyperref}

\DeclareCaptionStyle{ruled}{
    labelfont=normalfont,
    labelsep=colon,
    strut=off
} 

\title{Audio-Visual Dog Identification and Tracking in Real-World Videos}

\author{
    Aryahvishwa Babu\textsuperscript{\rm 1}\equalcontrib,
    Kingsley Jin\textsuperscript{\rm 2}\equalcontrib,
    Lillian Wang\textsuperscript{\rm 3}\equalcontrib,
    Tuan M. Dang\textsuperscript{\rm 4},\\
    Theron S. Wang\textsuperscript{\rm 4},
    Jeorge Joshi\textsuperscript{\rm 4},
    Annie Cao\textsuperscript{\rm 5},
    Kenny Q. Zhu\textsuperscript{\rm 4}\corresponding\thanks{This work is conducted at the National Science Foundation \textit{REU site: Animal Language Processing and Understanding} at University of Texas at Arlington and is supported by NSF Award No. 2349713.}
}

\affiliations{
    \textsuperscript{\rm 1}University of Maryland, College Park
    \textsuperscript{\rm 2}University of California, Merced
    \textsuperscript{\rm 3}Cornell University\\
    \textsuperscript{\rm 4}University of Texas at Arlington
    \textsuperscript{\rm 5}University of Washington\\
    ababu124@terpmail.umd.edu, kingsleyjin@ucmerced.edu, lw782@cornell.edu,\\
    \{txd0904,sxw7663,jxj5621\}@mavs.uta.edu, acao08@uw.edu, kenny.zhu@uta.edu
}

\begin{document}

\maketitle

\begin{abstract}
Identifying individual dogs from barks in real-world videos remains challenging
and largely unexplored. Solving this problem would enable long-term behavioral
analysis, veterinary monitoring, and animal welfare monitoring by attributing
vocalizations and behaviors to the correct individual. Real-world videos are
challenging because multiple dogs may bark simultaneously, become occluded, or
move in and out of view. Prior studies show that canine vocalizations and
visual appearance contain discriminative identity information, but these
modalities have largely been studied independently. We present a multimodal
dog-fingerprinting framework combining visual identification, multiple-object
tracking, and canine vocal fingerprinting to associate bark events with
individual dogs while maintaining persistent profiles. The framework
synchronizes bark events with tracked dogs and updates profiles using visual
appearance and vocal characteristics, enabling identification despite
occlusion, overlapping vocalizations, and off-screen barking. We also
introduce a benchmark of 46 real-world videos containing 213 manually annotated
bark events. Our framework correctly localizes the active visible dog in 61.83\% of visible
events at IoU $\geq 0.30$ and achieves a 91.33\% F1 score for visible versus
off-screen barking detection.
\end{abstract}

\section{Introduction}

Identifying individual dogs in videos supports animal behavior research,
veterinary monitoring, and long-term observation, where behaviors and
vocalizations must be attributed to the correct animal. Dog barks contain
individual-specific acoustic characteristics
\cite{yin2004barking,molnar2008classification,perezespinosa2018dog}, while
visual appearance supports dog recognition and animal re-identification
\cite{mougeot2019dogface,shinoda2024petface,williams2025animal}. These
modalities, however, have largely been studied separately.

\begin{figure}[ht]
    \centering
    \begin{subfigure}{\linewidth}
        \centering
        \includegraphics[width=\linewidth]{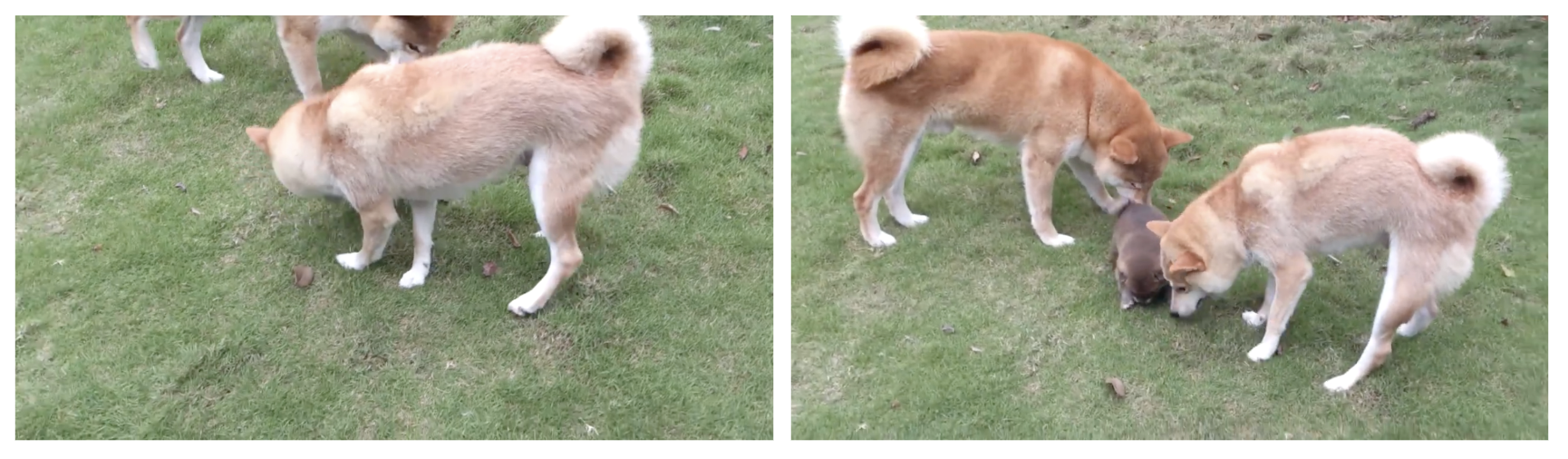}
        \caption{Occlusion}
    \end{subfigure}

    \begin{subfigure}{\linewidth}
        \centering
        \includegraphics[width=\linewidth]{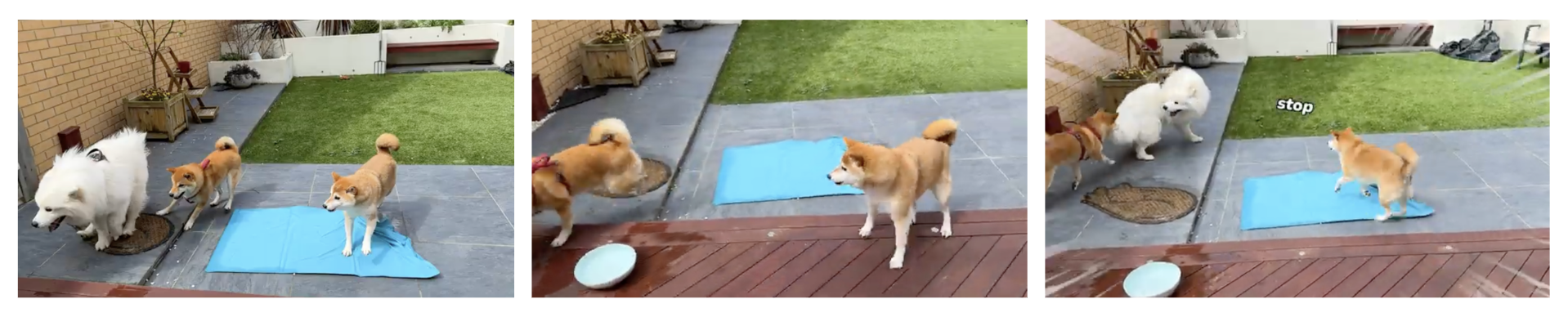}
        \caption{Exiting and re-entering}
    \end{subfigure}

    \caption{Challenges in dog identification under real-world conditions.}
    \label{fig:dog_challenges}
\end{figure}

We define \emph{multimodal dog fingerprinting} as associating bark events and
visual track fragments with persistent dog profiles throughout an
unconstrained video. The task extends audio-visual sound-source localization
\cite{senocak2018learning,huang2023egocentric} and active-speaker detection
\cite{roth2020ava} from event-level localization to identity association over
time. It must preserve identity across multiple barks, track fragmentation,
occlusion, disappearance, re-entry, and off-screen vocalization. Figure~\ref{fig:dog_challenges} highlights representative cases of these
challenging scenarios.

Real-world videos may contain simultaneous barks, environmental noise \cite{park2026towards},
occlusion, and dogs moving in and out of view. Audio identifies vocal
characteristics but cannot reliably localize a barking dog, particularly
during overlapping vocalizations. Visual tracking provides spatial and
temporal evidence but degrades under occlusion, identity switching, or
off-screen barking. Combining both modalities enables visible-barker
localization while preserving identity when either source is unavailable or
unreliable.

Existing canine datasets primarily support audio-only classification and
representation learning
\cite{abzaliev2024dog,li2025dog2vec,lekhak2025data,lekhak2025dogspeak,dang2025emotionalcanines, dang2026preview} or image-based
identification \cite{mougeot2019dogface,shinoda2024petface}. They do not jointly
annotate bark events, visible dog locations, off-screen vocalizations, and
persistent identities in real-world multi-dog videos. We address this gap
with a benchmark and framework integrating visual tracking, temporal
active-barker classification, vocal fingerprinting, and persistent profile
management. Our contributions are:

\begin{itemize}
    \item We formulate multimodal dog fingerprinting as associating bark events
    and visual track fragments with persistent dog profiles, extending
    audio-visual localization to identity association over time, and introduce a benchmark of 46 real-world videos with 213 annotated bark events, including 186 visible and 27 off-screen events, which we use to evaluate our method. The benchmark includes multiple-dog scenes, overlapping vocalizations,
occlusion, and re-identification\footnote{Our data and code is available at: \url{https://github.com/UTA-ACL2/dog_av_id_tracking}}.

    \item We propose a multimodal framework that achieves 61.83\% visible
    active-barker localization at IoU $\geq 0.30$, 91.33\% F1 for
    visible/off-screen detection, and 99.67\% profile-assignment coverage
    (303 of 304 detected bark events).

    \item We evaluate each component under realistic conditions. The temporal
    visual model achieves 73.3\% top-1 localization on multi-dog events, the
    hybrid audio fingerprint achieves 65.36\% open-set known-dog identification accuracy and 12.47\% EER using five enrollment barks, and persistent profile
    management merges 129 of 372 track fragments, reducing fragmentation by
    34.68\%.
\end{itemize}

\section{Related Work}

\subsection{Animal Tracking and Re-Identification}

Multiple-object tracking associates detections across frames to construct
trajectories. SORT uses motion prediction for frame-level association
\cite{bewley2016sort}, ByteTrack associates both high- and low-confidence
detections to recover partially occluded objects \cite{zhang2022bytetrack},
and BoT-SORT combines motion, appearance, and camera-motion compensation
\cite{aharon2022botsort}.

Animal tracking is complicated by similar appearances, deformation,
occlusion, and re-entry. \citet{williams2025animal} merged animal tracklets
using identity labels and temporal constraints. \citet{huang2023tracking}
introduced DETracker and the DogThruGlasses multi-dog benchmark, while
\citet{li2026fish} combined motion prediction, morphology-aware association,
and visual re-identification for similar-looking fish. Visual identity has
also been studied through dog-face verification and open-set recognition
\cite{mougeot2019dogface}, with PetFace providing a large-scale animal
face re-identification benchmark \cite{shinoda2024petface}. However, visual
evidence becomes unavailable when an animal is fully occluded or off-screen.

Related audio-visual work associates sounds with visible objects
\cite{senocak2018learning,huang2023egocentric} or identifies which visible person is speaking \cite{roth2020ava}. Active-barker localization must
additionally handle dog-specific motion, multiple candidates, and off-screen
vocalizations. Our framework links bark events and visual track fragments to
persistent dog identities over time.

\subsection{Canine Vocal Identity Recognition}

Dog vocalizations contain context- and individual-specific acoustic
characteristics \cite{yin2004barking,molnar2008classification}.
\citet{perezespinosa2018dog} used acoustic features to recognize individual
dogs, while \citet{abzaliev2024dog} evaluated pretrained human-speech
representations for canine identity and attribute classification. \citet{lekhak2025dogspeak} introduced DogSpeak with individual and breed
labels, while \citet{li2025dog2vec} learned canine-specific representations
from unconstrained online recordings.

Prior work generally treats canine identity recognition as classification
among dogs observed during training. Our audio component instead performs
enrollment-based open-set identification of unseen dogs by combining MFCCs
\cite{davis1980mfcc}, Dog2vec embeddings \cite{li2025dog2vec}, and learned
spectro-temporal representations.

\subsection{Audio-Based Identity Verification}

Human speaker-verification systems compare learned embeddings with enrolled
speaker profiles. ECAPA-TDNN combines channel attention, multi-scale temporal
features, and attentive statistics pooling \cite{desplanques2020ecapa}.
VoxBlink extends speaker verification to unconstrained audio-visual recordings
and uses face verification and active-speaker detection to curate
identity-labeled speech \cite{lin2024voxblink}.

Following this enrollment-based formulation, our model uses a ResNet encoder
\cite{he2016resnet} and ArcFace training \cite{deng2019arcface} to learn
compact canine vocal fingerprints. Unlike human speech, barks are brief and
vary across vocalization types and recording conditions. We combine these
fingerprints with visual tracks and active-barker predictions to maintain persistent dog profiles in real-world videos.

\begin{table*}[!t]
\centering
\small
\setlength{\tabcolsep}{5pt}
\begin{tabular}{lllr}
\hline
\textbf{Dataset} &
\textbf{Use} &
\textbf{Scale} &
\textbf{Samples} \\
\hline
Automated visual set
& Barker classifier: train
& 1,690 videos
& 12,738 windows \\

Tracking benchmark
& Visual tracking: test
& 21 videos
& 3,890 frames / 15,407 boxes \\

Multimodal benchmark
& Visual/fusion: test
& 46 usable videos
& 747 windows / 213 events \\

YouTube Channel Benchmark
& Audio fingerprint: train
& 426 dogs
& 415,361 clips \\

DogSpeak
& Open-set audio: test
& 150 dogs
& 65,263 clips \\
\hline
\end{tabular}
\caption{Datasets used to train and evaluate each framework component.}
\label{tab:dataset_summary}
\end{table*}

\section{Methods and Results}

Our framework combines visual tracking, active-barker classification, canine
audio fingerprinting, and identity-level fusion. Visual and vocal training
data were collected from unconstrained YouTube dog videos using the pipelines
of \citet{wang2024phonetic} and \citet{wang2025toward}.
Table~\ref{tab:dataset_summary} summarizes the dataset used for each task.

\begin{figure}[!ht]
    \centering
    \includegraphics[width=\columnwidth]{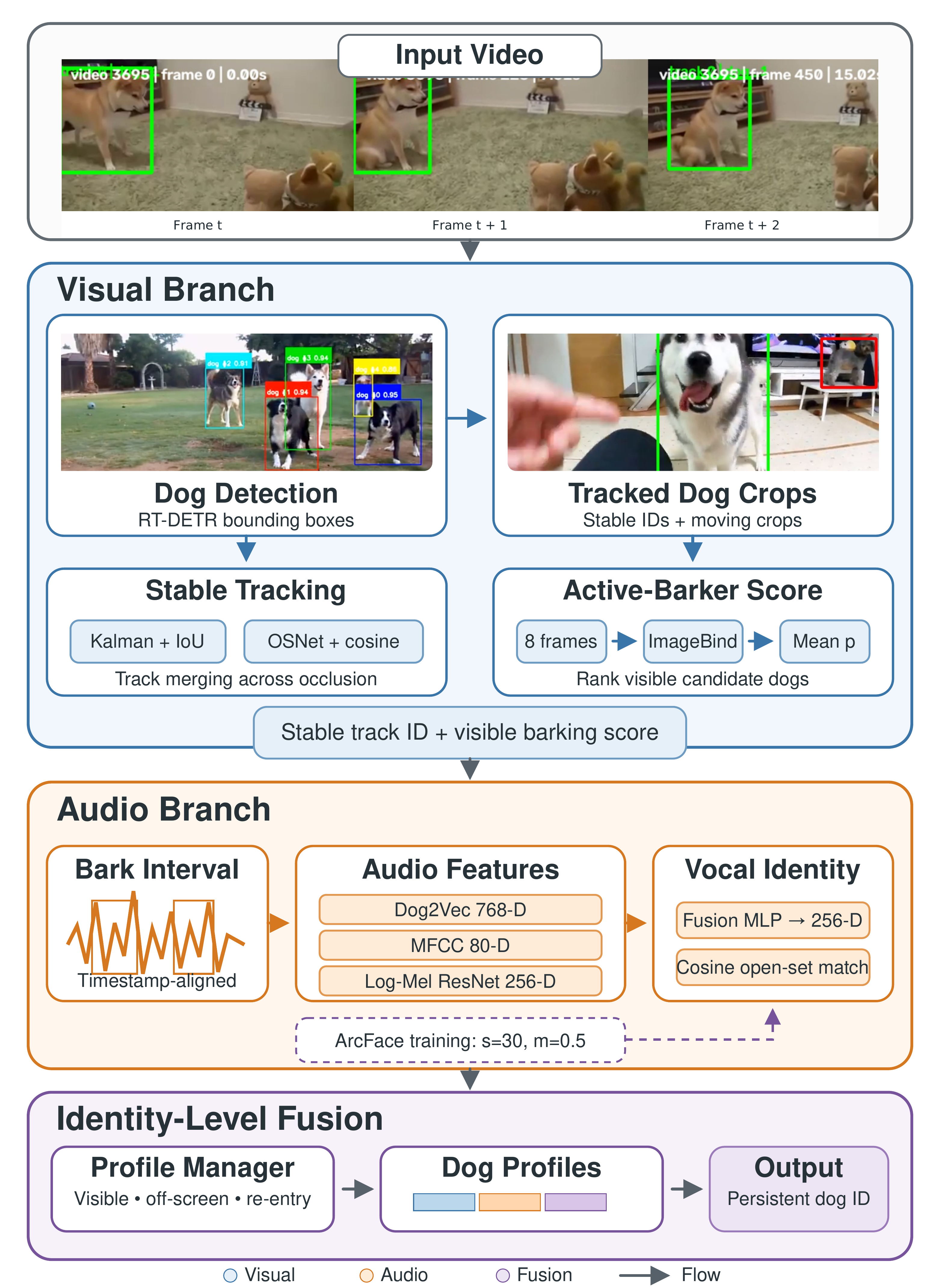}
    \caption{Overview of the proposed multimodal dog-fingerprinting
    framework. Visual tracking and active-barker classification are
    combined with open-set vocal fingerprinting to maintain persistent
    dog identities.}
    \label{fig:pipeline}
\end{figure}

Figure~\ref{fig:pipeline} summarizes the framework. Given a video and its bark
intervals, the visual branch tracks visible dogs and estimates each
candidate's barking probability. The audio branch extracts a fingerprint for
each bark and compares it with enrolled profiles. The fusion module combines
this evidence to maintain persistent dog identities.

All experiments are run on one NVIDIA RTX PRO 6000 Blackwell GPU.

\subsection{Visual Tracking}
\label{sec:visual_tracking}

The visual tracking component detects all visible dogs and maintains their
identities throughout each video. Detector identities may become fragmented
when dogs are occluded, temporarily leave the camera view, or later reappear.
Our framework therefore operates at two levels: short-term detection
association and long-term persistent profile management.

\subsubsection{Stable Detection Association}

We use the Ultralytics tracking wrapper with RT-DETR
\cite{zhao2024rtdetr} for frame-by-frame dog detection and short-term
tracking. The wrapper assigns temporary identities that may not persist across
occlusions or missed detections. We therefore introduce a stable identity layer
that maps temporary tracks to persistent identities.

Each stable identity uses a Kalman filter \cite{kalman1960filtering} to predict
subsequent bounding-box positions. We extract an appearance embedding for each
detected dog using the OSNet-x0.25 person re-identification architecture
\cite{zhou2019osnet}. Although designed for people, OSNet provides useful
embeddings for comparing dog crops. A weighted running average is maintained
for each stable identity.

When an unmatched detection appears, we compare it with all existing stable
identities using motion and appearance information. Motion similarity is
measured by intersection over union (IoU),
\(\operatorname{IoU}(B_1,B_2)=|B_1\cap B_2|/|B_1\cup B_2|\), between the
detected box and the Kalman-filter prediction. Appearance similarity is the
cosine similarity between the detection embedding and the aggregated
embedding of each stable identity.

The association score between detection \(d\) and stable identity \(i\) is

\begin{equation}
    S(d,i) =
    0.6\,S_{\mathrm{IoU}}(d,i)
    + 0.4\,S_{\mathrm{app}}(d,i),
\end{equation}
\noindent
where \(S_{\mathrm{IoU}}\) denotes IoU-based motion similarity and
\(S_{\mathrm{app}}\) denotes appearance cosine similarity. The detection is assigned to the highest-scoring stable identity when the
score exceeds an association threshold selected on development videos separate
from the tracking evaluation set; otherwise, it initializes a new identity.

\subsubsection{Persistent Profile Management}

Short-term association alone cannot fully address identity fragmentation after
long occlusions or extended absences. We therefore introduce a second
association layer that operates on completed track segments rather than
individual detections. This layer combines related track segments into
persistent dog profiles.

Candidate segments are compared using their aggregated OSNet-x0.25 appearance
embeddings together with temporal and spatial constraints. Non-overlapping
segments are merged when their appearance similarity exceeds 0.65. For
simultaneous tracks, duplicate detections are removed when their bounding-box
IoU exceeds 0.60 and their appearance similarity exceeds 0.45. Part-level duplicates require at least two common frames, appearance
similarity of at least 0.88, normalized center distance of at most 0.50, and
either IoU of at least 0.08 or intersection-over-smaller, defined as the intersection area divided by the
area of the smaller box, of at least 0.25. These thresholds were selected
empirically using development videos separate from the 21-video tracking
evaluation set.

Each persistent profile stores its associated track segments and aggregated
appearance representation. Consequently, a dog that disappears and later
re-enters the video can be re-associated with its previous profile rather than
receiving a new identity. 

\begin{table*}[!t]
\centering
\small
\setlength{\tabcolsep}{3pt}
\begin{tabular}{lcccccccc}
\hline
Tracker &
HOTA $\uparrow$ &
MOTA $\uparrow$ &
DetA $\uparrow$ &
AssA $\uparrow$ &
IDF1 $\uparrow$ &
FP $\downarrow$ &
FN $\downarrow$ &
IDs $\downarrow$ \\
\hline
BoT-SORT
& 42.61 & 26.78 & 40.14 & 47.10 & 40.52
& 3,447 & \textbf{5,718} & 75 \\

ByteTrack
& 41.44 & 27.11 & 40.16 & 44.68 & 36.25
& 3,367 & 5,758 & 74 \\

SORT
& 39.21 & 25.59 & 38.14 & 42.09 & 34.12
& \textbf{3,051} & 6,265 & 75 \\

StableTracker
& 46.85 & \textbf{27.34} & \textbf{40.38} & 56.44 & 46.42
& 3,368 & 5,768 & 34 \\

StableTracker + Profile Merging (Ours)
& \textbf{49.81} & 25.17 & 38.42 & \textbf{66.71}
& \textbf{51.11} & 3,221 & 6,211 & \textbf{11} \\
\hline
\end{tabular}
\caption{Tracking performance on the manually annotated benchmark. Best
results are shown in bold.}
\label{tab:tracking_results}
\end{table*}

\subsubsection{Evaluation}

We evaluate the tracking framework using 21 manually annotated videos
containing 3,890 annotated frames and 15,407 dog bounding boxes. We compare our
method with SORT \cite{bewley2016sort}, ByteTrack
\cite{zhang2022bytetrack}, and BoT-SORT \cite{aharon2022botsort}.

Performance is measured using higher-order tracking accuracy (HOTA), detection
accuracy (DetA), and association accuracy (AssA) \cite{luiten2021hota};
multiple-object tracking accuracy (MOTA), false positives (FP), false negatives
(FN), and identity switches (IDs) \cite{bernardin2008clear}; and IDF1
\cite{ristani2016performance}.

As shown in Table~\ref{tab:tracking_results}, StableTracker with persistent
profile merging achieves the highest HOTA, AssA, and IDF1. Profile merging
increases HOTA from 46.85 to 49.81 and AssA from 56.44 to 66.71 while reducing
identity switches from 34 to 11; the baselines produce 74--75 switches.
Duplicate suppression increases false negatives and slightly lowers MOTA
because some detections are removed, but substantially improves identity
consistency.

\subsection{Video-Based Bark Classification}
\label{sec:visual_bark_classification}

The video-based classifier assigns a \texttt{Barking} probability to each
visible tracked dog and ranks the candidates within each bark event. We train
it on a large weakly supervised dataset and evaluate it on a smaller manually
annotated benchmark.

\subsubsection{Manually Annotated Evaluation Dataset}

We construct a manually annotated benchmark for evaluating
visual active-barker localization. The original collection
contains 226 bark-centered clips from 46 usable source videos. After
preprocessing, the evaluation manifest contains 747 labeled
dog-track windows from 451 unique tracks across 213 bark
events. The difference between the original number of clips
and the final number of events results from preprocessing and
event validation.

Each video is manually reviewed to retain clips with
sufficiently clear dog visibility, reliable localization, and
interpretable barking behavior. The detected bounding-box
tracks are reviewed and corrected, and every track is assigned
a consistent identity within its corresponding event. Each
tracked dog is then labeled as either \texttt{Barking} or
\texttt{Not\_Barking} according to the manually annotated bark interval. The final
benchmark contains 343 \texttt{Barking} windows and 404
\texttt{Not\_Barking} windows.

For event-level evaluation, we retain events containing
exactly one valid \texttt{Barking} candidate. Of the 213 bark events,
143 satisfy this requirement. This set contains 53
single-candidate events and 90 multi-candidate events. The
multi-candidate subset directly evaluates the primary visual
task: selecting the correct barking dog when multiple dogs are
simultaneously visible. The remaining 70 events are excluded
only from event-level top-1 evaluation because they do not
contain exactly one valid \texttt{Barking} candidate; their labeled
windows can still be used for window-level evaluation.

\subsubsection{Automated Training Dataset}

To obtain additional training data without exhaustive manual
annotation, we develop an automated dataset-generation
pipeline. The pipeline identifies videos containing
approximately one visible dog, allowing provided bark
timestamps to be used as weak supervision. For each source
video, we calculate the average number of tracked dog
identities and retain videos with an average identity count of
at most 1.1. When approximately one dog is visible, a provided
bark timestamp can be associated with that dog with
substantially less ambiguity. 

\texttt{Barking} samples use the exact provided start and end
timestamps without additional temporal padding. \texttt{Not\_Barking}
samples are extracted as two-second windows from temporal
regions that do not overlap any provided bark interval. We
retain at most 10 \texttt{Not\_Barking} windows from each source video
and remove samples shorter than 0.5 seconds.

Dogs are first tracked throughout each source video, after
which interval-specific moving crops are generated from the
track data. The same detector, tracker, bounding-box
interpolation procedure, spatial transformation, and output
format are used for both classes to avoid introducing
class-specific artifacts. No spatial padding is added to the
bounding boxes, keeping each crop tightly aligned with the
tracked dog. Every generated sample contains a dog-centered
video crop, aligned audio, a class label, its source-video
identifier, and associated metadata.

Before quality control, the pipeline generates 15,871 samples
from 2,109 source videos, including 5,214 \texttt{Barking} and 10,657
\texttt{Not\_Barking} samples. After filtering, audio-based quality
control, and removal of overlap with previously processed
data, the final training dataset contains 12,738 samples from
1,690 source videos, comprising 5,210 \texttt{Barking} and 7,528
\texttt{Not\_Barking} samples.

The dataset is split by source video into 10,190 training
samples and 2,548 validation samples (80\%/20\%). This group-based split prevents windows from the same recording from appearing
in both partitions, reducing leakage from shared frames and recording
conditions. The manually annotated
benchmark is excluded from training and retained exclusively
for evaluation.

\subsubsection{Audio-Based Quality Control}

Weakly supervised extraction of \texttt{Not\_Barking} samples can
introduce label noise when unannotated barks or other dog
vocalizations occur outside the provided bark intervals. To
reduce this noise, we apply a PANNs-based audio classifier
\cite{kong2020panns} as a quality-control filter.

The classifier measures evidence for direct barking sounds,
other dog vocalizations, general dog-related sounds, and
broader animal sounds. We marked a \texttt{Not\_Barking} sample as suspicious if its PANNs score was at least 0.18 for bark-like sounds, 0.15 for other dog vocalizations, 0.20 for general dog-related sounds, or 0.10 for broader animal sounds. Samples that did not meet any of these thresholds were retained as clean candidates.
This audio signal is used only for training-data quality
control and is not used to predict the identity of the
barking dog. The manually annotated visual dataset remains
the primary benchmark for evaluating active-barker
localization.

\subsubsection{ImageBind Visual Classification}

We use ImageBind as a frozen visual feature extractor to
classify tracked dog clips as \texttt{Barking} or
\texttt{Not\_Barking}
\cite{girdhar2023imagebind}. Only its visual encoder is used,
and its pretrained parameters remain fixed during training.
For each dog-track window, the moving crop produced by the
tracking pipeline is converted to the ImageBind visual input
format and passed through the frozen encoder. The encoder
produces a 1,024-dimensional visual representation.

The extracted embeddings are standardized using a
StandardScaler and passed to a logistic-regression classifier
implemented using scikit-learn
\cite{pedregosa2011scikit}. The classifier outputs the
probability that the visible dog is barking. Only the
logistic-regression classifier is trained, and class weighting
is used to account for the unequal numbers of \texttt{Barking} and
\texttt{Not\_Barking} samples.

During inference, each candidate dog is processed
independently. If a track contains multiple windows, its
window probabilities are averaged to obtain a track-level
score. Within each bark event, the candidate tracks are ranked
by their scores, and the highest-scoring track is selected as
the predicted active barker.

\subsubsection{Temporal Probability Aggregation}

A representation extracted from a small number of frames may
miss brief mouth, head, or body movements associated with
barking. To increase temporal coverage without fine-tuning the
ImageBind encoder, we uniformly sample eight frames from each
dog-track window and divide them into four chronological
two-frame pairs.

Each pair is independently processed by the same frozen
ImageBind encoder and trained logistic-regression classifier.
For candidate track \(i\), this produces four \texttt{Barking}
probabilities \(p_{i,1},\ldots,p_{i,4}\). The final visual
\texttt{Barking} score is calculated as

\begin{equation}
    p_i^{v} =
    \frac{1}{4}\sum_{j=1}^{4}p_{i,j}.
\end{equation}
\noindent
For a bark event containing a set of visible candidate tracks
\(\mathcal{C}_t\), the predicted active barker is

\begin{equation}
    \hat{i}_t^{v}
    =
    \arg\max_{i \in \mathcal{C}_t} p_i^{v}.
\end{equation}
\noindent
This approach increases the probability of observing
brief barking-related motion without adding trainable encoder
parameters.

We additionally evaluate a twelve-frame sampling variant and
two learned aggregation models. The temporal statistics head
combines summary statistics calculated from the four
pair-level predictions, while the temporal attention head
learns weights for the pair-level representations. These
models test whether learned temporal weighting improves upon
simple probability averaging.

All temporal variants use the same frozen ImageBind visual
encoder. The eight- and twelve-frame mean variants reuse the
ImageBind+LR classifier across multiple frame pairs and average
the resulting probabilities, while the temporal statistics and
temporal attention variants use learned aggregation heads over
the temporal outputs.

\subsubsection{Evaluation}

We evaluate visual classification at three levels.
Window-level metrics measure binary classification of
individual dog-track windows. Track-level metrics average
predictions from multiple windows belonging to the same
track. Event-level top-1 accuracy measures whether the true
barking dog receives the highest score among all visible
candidates.

\begin{table*}[!t]
\centering
\small
\setlength{\tabcolsep}{3pt}
\begin{tabular}{lccccc}
\hline
Method &
Window Acc. $\uparrow$ &
Window AUC $\uparrow$ &
Track Acc. $\uparrow$ &
Track AUC $\uparrow$ &
Multi-dog Top-1 $\uparrow$ \\
\hline
Zero-shot ImageBind       & 45.4 & 0.487 & 43.7 & 0.472 & 47.8 \\
ImageBind + LR            & 56.8 & 0.572 & 59.2 & 0.588 & 53.3 \\
8-frame mean              & 58.6 & \textbf{0.594} & \textbf{61.9} &
\textbf{0.626} & \textbf{73.3} \\
12-frame mean             & \textbf{58.8} & 0.581 & \textbf{61.9} &
0.619 & 65.6 \\
Temporal statistics       & 58.1 & 0.561 & 61.2 & 0.590 & 65.6 \\
Temporal attention        & 50.5 & 0.507 & 50.3 & 0.556 & 50.0 \\
\hline
\end{tabular}
\caption{Visual barking-classification results on the manually annotated
benchmark. Multi-dog top-1 accuracy uses 90 multi-candidate events.}
\label{tab:visual_classification_results}
\end{table*}

Accuracy values in Table~\ref{tab:visual_classification_results}
are reported as percentages, while ROC-AUC is reported on a scale
from zero to one. The event-specific random expectation for the
multi-candidate subset is 41.8\%.

Zero-shot ImageBind performs near chance (45.4\% accuracy and 0.487 AUC).
Training logistic regression improves these results to 56.8\% and 0.572,
respectively, and reaches 53.3\% multi-dog top-1 accuracy. Eight-frame
averaging provides the best multi-dog localization, improving top-1 accuracy
to 73.3\% while achieving 58.6\% window accuracy and the highest window AUC
of 0.594.

Twelve-frame averaging achieves the highest window accuracy and ties the
eight-frame model for the highest track accuracy, but its multi-dog accuracy
is lower at 65.6\%. Its 78.3\% overall event accuracy also
includes 53 single-candidate events that are selected automatically. Additional
frames may therefore improve binary classification without improving relative
ranking among simultaneously visible dogs.

The learned temporal heads generalize less effectively. Although the
statistics and attention heads achieve validation AUCs of 0.712 and 0.722 on
the automated set, their manual-benchmark AUCs fall to 0.561 and 0.507. This
distribution shift favors the simpler eight-frame average, which we use in the
complete framework.

\subsection{Audio Fingerprinting}
\label{sec:audio_fingerprinting}

The audio component maps canine vocalizations to identity embeddings compared
across recordings using cosine similarity. It combines spectral and learned
representations with enrollment-based identification, allowing unseen dogs to
be enrolled using few barks.

We train on the YouTube Channel Benchmark, constructed from videos collected
by \citet{wang2024phonetic}. We retain channels containing at least ten videos
of the same identifiable dog and follow the pipeline of
\citet{wang2024phonetic,wang2025toward} to extract canine vocal segments. The
dataset contains 426 dog identities and 415,361 bark clips with variation in
background noise, recording environment, and microphone quality. All clips
are used to train the audio-fingerprinting model.

Evaluation uses DogSpeak \cite{lekhak2025dogspeak}, containing 156 identity
labels and 77,202 bark clips across five breeds. Dog~7 is divided into
identities 7a and 7b because of its unusually large number of clips. After
clip-count filtering, the evaluation set contains 150 identities and 65,263
clips. No DogSpeak data are used for training, ensuring identity-disjoint
evaluation.

\begin{table*}[!t]
\centering
\small
\setlength{\tabcolsep}{4pt}
\begin{tabular}{llccccc}
\hline
Method & Features & Acc. $\uparrow$ & F1 $\uparrow$ &
EER $\downarrow$ & minDCF $\downarrow$ & DIR@1\% $\uparrow$ \\
\hline
MFCC & MFCC & 15.73 & 6.81 & 37.90 & 0.0100 & 0.33 \\
Dog2vec & Dog2vec & 28.10 & 11.99 & 26.09 & 0.0095 & 5.00 \\
Feature Fusion & MFCC + Dog2vec & 28.25 & 12.24 & 27.24 & 0.0095 &
4.89 \\
ResNet-TSP & 80-d log-Mel & 56.89 & 32.71 & 14.33 & 0.0073 & 22.30 \\
\textbf{Ours} & Hybrid embedding & \textbf{65.36} & \textbf{37.14} &
\textbf{12.47} & \textbf{0.0063} & \textbf{29.48} \\
\hline
\end{tabular}
\caption{Audio-representation comparison on DogSpeak using five enrollment
barks. Accuracy, macro F1, EER, and DIR are percentages; DIR@1\% is measured
at a 1\% false-accept rate.}
\label{tab:audio_results}
\end{table*}

\subsubsection{Audio Fingerprint Learning}

Each bark is represented using three complementary features. We extract an
80-dimensional MFCC representation \cite{davis1980mfcc}, a 768-dimensional
embedding from the pretrained Dog2vec model \cite{li2025dog2vec}, and an
80-channel log-Mel spectrogram. A lightweight ResNet-TSP encoder
\cite{he2016resnet}, inspired by VoxBlink \cite{lin2024voxblink}, maps the
log-Mel spectrogram to a 256-dimensional spectro-temporal embedding.

The three representations are concatenated into a 1,104-dimensional vector and
passed through an MLP with layer dimensions 1,024, 512, and 256. Each hidden
layer uses batch normalization, ReLU activation, and dropout of 0.2. The output
is L2-normalized to produce a 256-dimensional audio fingerprint.

We train the network for 30 epochs using ArcFace
\cite{deng2019arcface}, with scale 30 and angular margin 0.5. Optimization uses
AdamW \cite{loshchilov2019adamw}, a batch size of 512, and a learning rate of
\(10^{-3}\). The ArcFace classification head is discarded after training,
leaving only the embedding network for inference.

\subsubsection{Enrollment-Based Identification}

Following enrollment-based speaker-verification protocols
\cite{desplanques2020ecapa,lin2024voxblink}, we embed and average
\(K\in\{1,3,5\}\) barks from dog \(i\) to construct its profile:

\begin{equation}
    \mathbf{P}_i =
    \frac{1}{K}\sum_{k=1}^{K} f_{\theta}(x_{i,k}),
\end{equation}

\noindent
where \(f_{\theta}\) is the embedding network and \(x_{i,k}\) is an enrollment
bark. We restrict evaluation to five or fewer barks to follow a low-enrollment
speaker-verification setting and because the multimodal framework must form
profiles from only a few initial vocalizations.

Given probe embedding \(\mathbf{z}\), its similarity to profile \(i\) is

\begin{equation}
    s_i =
    \frac{\mathbf{z}^{\top}\mathbf{P}_i}
    {\lVert\mathbf{z}\rVert_2\lVert\mathbf{P}_i\rVert_2}.
\end{equation}
\noindent
The highest-scoring identity is selected when its similarity exceeds a
rejection threshold; otherwise, the probe is treated as unknown. The complete framework uses an audio-similarity threshold of 0.60, selected
using a separate development gallery.

For evaluation, we construct five random splits of 120 enrolled identities and
30 unknown identities. Enrollment clips are excluded from the known probe set.
Known-dog accuracy and macro F1 evaluate identification among enrolled
identities. EER is obtained by sweeping the similarity threshold over known and
unknown probes. DIR is reported at false-accept rates (FARs) of 1\%, 5\%, and
10\%, with a separate operating threshold selected for each target FAR. We
also report minDCF with a target prior of 0.01. All results are averaged over
the five splits.

\subsubsection{Evaluation}

Table~\ref{tab:audio_results} shows that the hybrid model performs best,
achieving 65.36\% known-dog accuracy, 12.47\% EER, and 29.48\% DIR@1\% with
five enrollment barks. Relative to ResNet-TSP, it improves accuracy and
DIR@1\% by 8.47 and 7.18 percentage points.
Table~\ref{tab:audio_enrollment_results} shows that increasing enrollment
from one to five barks raises accuracy from 40.69\% to 65.36\% and reduces
EER from 20.38\% to 12.47\%.

\begin{table}[t]
\centering
\small
\setlength{\tabcolsep}{2pt}
\begin{tabular}{cccccc}
\hline
Barks & Acc. $\uparrow$ & F1 $\uparrow$ & EER $\downarrow$ &
minDCF $\downarrow$ & DIR@1\% $\uparrow$ \\
\hline
1 & 40.69 & 21.13 & 20.38 & 0.0084 & 11.49 \\
3 & 59.23 & 32.60 & 14.32 & 0.0069 & 23.59 \\
5 & \textbf{65.36} & \textbf{37.14} & \textbf{12.47} &
\textbf{0.0063} & \textbf{29.48} \\
\hline
\end{tabular}
\caption{Effect of enrollment size on DogSpeak. Accuracy, macro F1, EER, and
DIR are percentages; DIR@1\% is measured at a 1\% false-accept rate.}
\label{tab:audio_enrollment_results}
\end{table}

\subsection{Multimodal Fusion Framework}
\label{sec:multimodal_fusion}

The fusion framework combines persistent visual profiles, active-barker
predictions, and vocal fingerprints. For each bark event, the visual classifier
selects a candidate profile, while the audio model compares the bark embedding
with existing vocal profiles. These signals are fused at the identity level
rather than by averaging model predictions.

\subsubsection{Evaluation Data}

We evaluate the framework on 46 unconstrained YouTube videos containing 213
annotated bark events: 186 visible and 27 off-screen. Visible events identify
the barking dog, whereas off-screen events indicate that no barking dog is
visible. These annotations support active-dog localization and
visible/off-screen detection.

\subsubsection{Multimodal Identity Fusion}
\label{sec:identity_fusion}

The framework maintains a dynamic set of hypothesized dog profiles:

\begin{equation}
    \mathcal{P}_i =
    \left(\mathbf{v}_i,\mathbf{a}_i,n_i^v,n_i^a\right),
\end{equation}
\noindent
where \(\mathbf{v}_i\) and \(\mathbf{a}_i\) are aggregated visual and audio
representations, and \(n_i^v\) and \(n_i^a\) are their observation counts.
Temporary tracks are mapped to persistent profiles. When the visual classifier
identifies a visible barker, the event is associated with its visual profile.

For the same event, the audio model produces fingerprint \(\mathbf{a}_t\),
whose similarity to profile \(i\) is

\begin{equation}
    s_{i,t}^{a} =
    \frac{\mathbf{a}_t^{\top}\mathbf{a}_i}
    {\lVert\mathbf{a}_t\rVert_2\lVert\mathbf{a}_i\rVert_2}.
\end{equation}
\noindent
The highest-scoring match is accepted when its similarity is at least
\(\tau_a=0.60\). Agreement between visual and audio evidence reinforces the
profile. An accepted audio observation updates its mean incrementally:
\noindent
\begin{equation}
    \bar{\mathbf{a}}_i^{\,\prime} =
    \frac{n_i^a\bar{\mathbf{a}}_i+\mathbf{a}_t}
         {n_i^a+1},
\end{equation}
\noindent
after which it is L2-normalized. Visual representations are aggregated
similarly.

When the modalities select different profiles, the system records a conflict
rather than merging them. Off-screen assignment relies on existing audio
profiles, while audio observations below the threshold remain unmatched and
can initialize a new profile.

\subsubsection{Evaluation}
\label{sec:fusion_results}

For each visible event, IoU measures overlap between the predicted barking-dog
box and its annotation. Localization accuracy is the percentage of events with
IoU at least 0.30, while mean IoU measures average spatial overlap. Treating
visible barking as the positive class, we report precision, recall, and their
harmonic mean, F1.

As shown in Table~\ref{tab:fusion_results}, the system correctly localizes
61.83\% of the 186 visible events, with a mean spatial IoU of 0.589.
Visible/off-screen detection achieves 86.89\% precision, 96.24\% recall, and
91.33\% F1.

\begin{table}[t]
\centering
\small
\setlength{\tabcolsep}{3pt}
\begin{tabular}{lr}
\toprule
\textbf{Metric} & \textbf{Result} \\
\midrule
Correct localization at IoU $\geq 0.30$ & 61.83\% \\
Mean spatial IoU & 0.589 \\
Visible/off-screen precision & 86.89\% \\
Visible/off-screen recall & 96.24\% \\
Visible/off-screen F1 & 91.33\% \\
Profile-assignment coverage & 99.67\% \\
Track-fragmentation reduction & 34.68\% \\
\bottomrule
\end{tabular}
\caption{Performance of the complete multimodal framework.}
\label{tab:fusion_results}
\end{table}

The 213 manually validated events support localization and visible/off-screen
evaluation. For coverage analysis, we process all 304 timestamped intervals
from the same 46 videos, including those without spatial annotations. The
system assigns 303 events to persistent profiles (99.67\% coverage), measuring
assignment availability rather than identity correctness. The profile manager
also consolidates 372 temporary IDs into 243 profiles, reducing fragmentation
by 129 IDs (34.68\%).

No existing end-to-end baseline jointly performs persistent dog tracking,
active-barker localization, open-set vocal identification, and multimodal
profile management. We therefore use task-specific comparisons: SORT,
ByteTrack, and BoT-SORT for tracking; spatial and temporal ImageBind variants
for active-barker classification; and MFCC, Dog2Vec, feature-fusion, and
ResNet-TSP audio baselines. The fusion results characterize complete-system
behavior rather than superiority over an existing fusion method.

Manual inspection reveals recurring localization failures involving limited visibility and non-frontal viewpoints. Of the 71 localization errors, 41 (57.7\%) involve limited visibility, and 43 (60.6\%) involve non-frontal viewpoints. Limited visibility includes cropping or partial/severe occlusion of the barking dog, while non-frontal viewpoints include sideways, backward, or mixed views. These categories often overlap. However, 30 errors exhibit no annotated visibility issue, suggesting that other factors also contribute, including poor video quality, multiple visible dogs, and tracking errors.

The broader pipeline remains sensitive to detection and tracking errors.
Separate head and body detections create duplicate tracks, while one box
containing multiple dogs prevents profile separation. Static images and sparse
tracks can also create uncertain profiles. Temporal re-identification,
duplicate suppression, internal-motion analysis, and part-based merging
mitigate these errors but cannot recover tracks containing mixed identities.
Noisy or overlapping vocalizations can further weaken audio evidence and
produce cross-modal conflicts.
\section{Conclusion}
\label{sec:conclusion}

We presented a multimodal framework for associating bark events with
persistent dog identities in real-world videos. On our 46-video benchmark,
the framework correctly localizes 61.83\% of visible barking events at IoU
$\geq 0.30$ and achieves 91.33\% F1 for visible/off-screen detection. In a
separate pipeline-level coverage analysis on the same videos, it assigns 303
of 304 processed bark events to persistent profiles (99.67\% coverage).
Remaining challenges include overlapping vocalizations, severe occlusion,
noisy audio, limited enrollment, and visually or acoustically similar dogs.
Future work will expand the benchmark and investigate jointly learned
audio-visual representations.

\bibliography{aaai2027}

\end{document}